%% file: neurips_2026.tex
\documentclass{article}

 \usepackage[main, final]{neurips_2026}

\usepackage[utf8]{inputenc} %
\usepackage[T1]{fontenc}    %
\usepackage{hyperref}       %
\usepackage{url}            %
\usepackage{booktabs}       %
\usepackage{amsfonts}       %
\usepackage{nicefrac}       %
\usepackage{microtype}      %
\usepackage{xcolor}         %
\usepackage{amsmath}        %
\usepackage{tikz}
\usetikzlibrary{arrows.meta,positioning}
\usepackage{needspace}
\hypersetup{hidelinks,pdftitle={Scaling Laws for SAE Training Data},pdfauthor={Nikita Koriagin, Nikita Balagansky, Daniil Gavrilov}}
\usepackage{graphicx}       %

\title{Scaling Laws for SAE Training Data}

\author{%
   Nikita Koriagin\thanks{Work done while at T-Tech.} \\
   Yandex \\
   \And
   Nikita Balagansky\\
   T-Tech \\
   \And
   Daniil Gavrilov \\
   T-Tech \\
}

\begin{document}

\maketitle

\begin{abstract}
Activation storage is often a bottleneck in sparse autoencoder (SAE) training. We study how many unique activations an SAE needs when its total training budget is fixed. Quality improves rapidly as the buffer grows, then reaches a plateau where replaying the buffer is nearly as effective as using fresh activations throughout training. Beyond this point, additional unique activations bring little benefit to reconstruction or interpretability, so the buffer need not grow with the training budget. We describe this tradeoff between diversity and repetition with a scaling law for limited data. We validate it across dictionary sizes and token budgets, early, middle, and late layers of Llama-3.1, Qwen3 model sizes, and BatchTopK, TopK, and JumpReLU SAEs. Sweeps using mathematical text and Python code test the effect of changing the activation source. Evaluation covers explained variance (EV), reconstruction MSE, feature absorption, automated interpretability, and sparse probing. These results give a practical training rule: choose the smallest buffer that reaches the quality plateau and replay it for the remaining budget. In our experiments, this reduces activation storage by $8\text{--}64\times$ with negligible quality loss.

\end{abstract}

\section{Introduction}

Sparse autoencoders (SAEs) decompose neural representations into interpretable features~\citep{cunningham2023sparse,gao2024scaling,lieberum2024gemma}. Training them at scale can require several terabytes of stored activations. Generating activations once lets different dictionary sizes, sparsities, and architectures share the cost of forward passes through the base model. Storage still grows with the training budget if every SAE update uses fresh activations. We separate the total number of activation presentations to the optimizer from the number of \emph{unique} activations in the buffer, and ask how large that buffer must be before replay can replace further fresh samples.

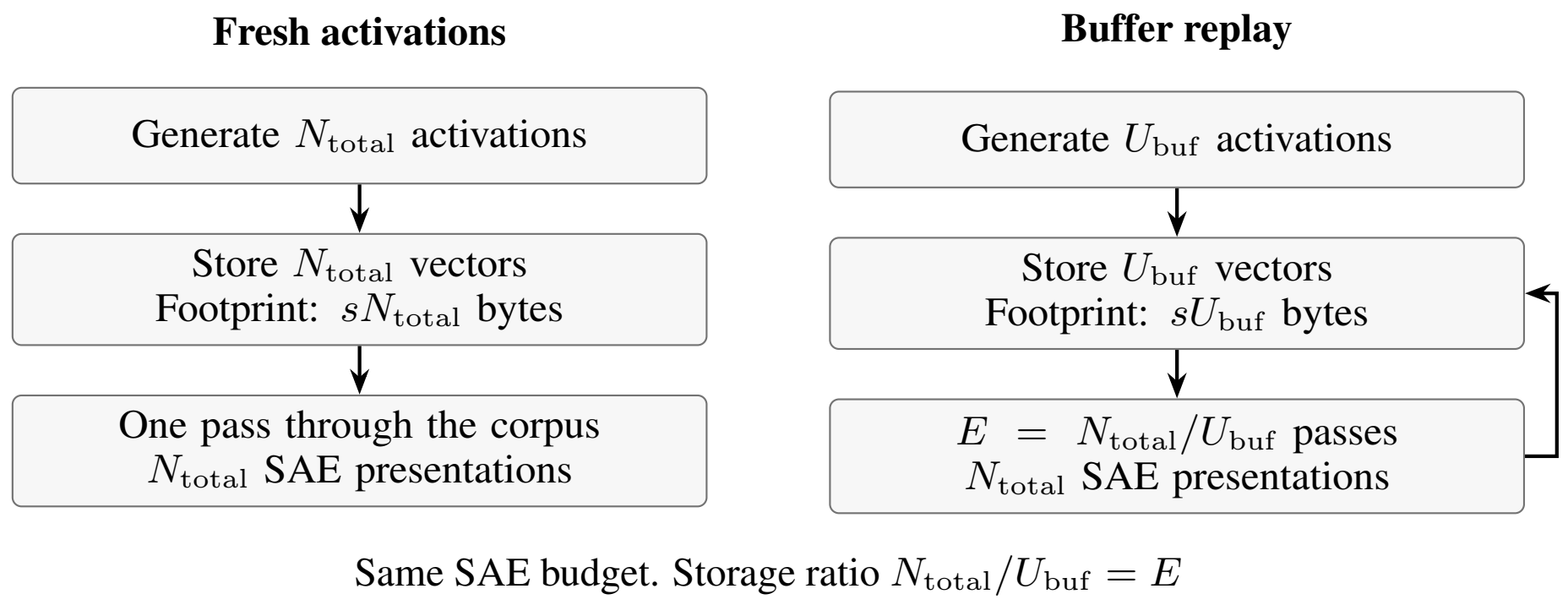
\begin{figure}[t]
\centering
\input{figures/replay_pipeline.tex}
\caption{Fresh activations and replay at a fixed SAE optimization budget. The replay pipeline generates and stores $U_{\mathrm{buf}}$ activations once and uses $E=N_{\mathrm{total}}/U_{\mathrm{buf}}$ passes. $s$ denotes bytes per activation. Generation counts refer to token activations. Batching and sequence packing determine the number of calls to the base model.}
\label{fig:pipelines}
\end{figure}

To learn a feature's direction and threshold, the autoencoder must encounter that feature in its training activations. If samples were independent, a feature with occurrence probability $p$ would be absent from a buffer of size $U$ with probability $(1-p)^U$. Replay gives the optimizer more opportunities to fit features already represented in the buffer. It cannot introduce an absent feature. The number of unique activations measures distributional coverage, while total presentations measure optimization effort. Contextual activations are correlated, so this argument motivates our controlled sweep without providing a guarantee on sample complexity. Figure~\ref{fig:pipelines} shows the two training schedules.

We vary buffer diversity across dictionary sizes and budgets while holding total optimization tokens fixed within each sweep. Reconstruction improves rapidly as small buffers grow and then plateaus. A scaling law using effective activation counts, adapted from \citet{muennighoff2023scaling}, fits the 96-run grid with RMSE $4.8\times10^{-3}$ EV. The Akaike Information Criterion (AIC) favors a version whose repetition scale depends on dictionary size. The fitted law also interpolates held-out budgets and predicts held-out widths when the dictionaries are sufficiently trained.

Saturation also appears across Llama layers and Qwen model sizes, with different SAE gates and with activations from mathematical text and Python code. SAEBench evaluations show strong AutoInterp and reconstruction performance with the buffer chosen from the EV plateau. Feature absorption and sparse probing provide further checks on the consequences of replay. We use the EV plateau to select a buffer and replay it for the remaining training budget. The resulting buffer sizes reduce activation storage by $8\text{--}64\times$ in our grid. Tolerance tests quantify how these storage savings change with the target EV gap.

\begin{figure}[t]
    \centering
    \includegraphics[width=\linewidth]{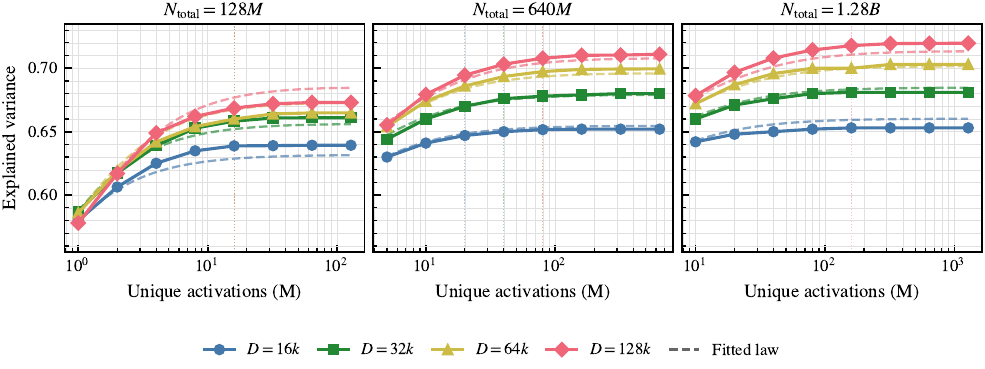}
    \caption{
    Explained variance as a function of unique activation count under fixed total SAE training budgets. With total optimization tokens fixed in each panel, moving right increases activation diversity and decreases the number of epochs over the buffer. EV improves sharply when activation diversity is low and then saturates.
    }
    \label{fig:ev_vs_unique}
\end{figure}

\section{Related Work}

\textbf{Sparse autoencoders for mechanistic interpretability.}
\citet{cunningham2023sparse} trained sparse autoencoders to decompose language model activations into sparse, interpretable features. \citet{gao2024scaling} studied SAE scaling across sparsities, data budgets, and layers. \citet{templeton2024scaling} trained SAEs on Claude 3 Sonnet. \citet{lieberum2024gemma} released Gemma Scope, a suite of open SAEs spanning layers and dictionary widths in Gemma 2. \citet{bills2023language} and \citet{paulo2024automatically} study automated feature interpretation, while SAEBench~\citep{karvonen2024saebench} evaluates SAE quality on downstream tasks relevant to interpretability. We study unique activations and total training presentations as separate variables. BatchTopK~\citep{bussmann2024batchtopk} and JumpReLU~\citep{rajamanoharan2024jumping} change the tradeoff between sparsity and reconstruction. \citet{chanin2025sparse} discuss how inappropriate $L_0$ can mix features. Our law therefore holds sparsity fixed, and the JumpReLU comparison tests saturation without establishing optimal sparsity or an architecture ranking at matched $L_0$.

\textbf{Scaling laws for limited data and compute.}
\citet{hoffmann2022training} derived how to allocate compute between language model size and training tokens (Chinchilla). \citet{muennighoff2023scaling} extended this framework to finite text corpora, where training must reuse data. Their effective token count describes the diminishing returns from repetition and supports scaling laws when data are limited. Other scaling studies distinguish the capacity of routed models from compute~\citep{clark2022unified} and examine data quality and training strategy~\citep{chen2025revisiting}. We adapt this framework of effective tokens to SAE training to model the joint effects of activation diversity and dictionary width.
\section{Setup}

\textbf{Base models and activations.}
The main experiments use residual stream activations from layer 15 of Llama-3.1 8B. We sample sequences from FineWeb~\citep{penedo2024fineweb} and construct activation replay buffers at a fixed layer. We also validate on layers 5 and 25 of Llama-3.1 8B and on comparable layers near the middle of Qwen3-1.7B, Qwen3-4B, and Qwen3-8B.

\textbf{SAE architecture.}
The main grid trains BatchTopK sparse autoencoders~\citep{bussmann2024batchtopk} with target average sparsity $k=32$ and dictionary sizes $D \in \{16\mathrm{k},32\mathrm{k},64\mathrm{k},128\mathrm{k}\}$. We test whether the observed saturation pattern transfers beyond this setting using BatchTopK for validation across layers, standard TopK~\citep{gao2024scaling} for Qwen validation, and comparisons with JumpReLU~\citep{rajamanoharan2024jumping} at $D=32\mathrm{k}$.

\textbf{Diversity--repetition grid.}
We train with total token budgets $N_{\mathrm{total}} \in \{128\mathrm{M},640\mathrm{M},1.28\mathrm{B}\}$. For each budget, we vary the number of unique activations $U$ by sweeping
\[
U/N_{\mathrm{total}} \in \{1/128,1/64,1/32,1/16,1/8,1/4,1/2,1\}.
\]
Thus each curve holds total optimization tokens fixed while moving from training with fresh activations throughout ($U=N_{\mathrm{total}}$) to 128 replay epochs over a small buffer ($U=N_{\mathrm{total}}/128$). This isolates activation diversity from total optimization compute.

\begin{table}[t]
\centering
\caption{
Saturation intervals for each dictionary size and total training budget ($\epsilon = 0.005$).
}
\label{tab:saturation_intervals}
\begin{tabular}{lrrrr}
\toprule
Budget & Dict.\ size & Plateau EV & Saturation interval & Storage reduction \\
\midrule
128M  & 16k  & $0.6393$ & 8M--16M     & 8--16$\times$ \\
128M  & 32k  & $0.6610$ & 16M--32M    & 4--8$\times$ \\
128M  & 64k  & $0.6648$ & 16M--32M    & 4--8$\times$ \\
128M  & 128k & $0.6730$ & 16M--32M    & 4--8$\times$ \\
\midrule
640M  & 16k  & $0.6522$ & 20M--40M    & 16--32$\times$ \\
640M  & 32k  & $0.6803$ & 40M--80M    & 8--16$\times$ \\
640M  & 64k  & $0.6995$ & 80M--160M   & 4--8$\times$ \\
640M  & 128k & $0.7110$ & 80M--160M   & 4--8$\times$ \\
\midrule
1.28B & 16k  & $0.6530$ & 20M--40M    & 32--64$\times$ \\
1.28B & 32k  & $0.6810$ & 40M--80M    & 16--32$\times$ \\
1.28B & 64k  & $0.7030$ & 80M--160M   & 8--16$\times$ \\
1.28B & 128k & $0.7197$ & 160M--320M  & 4--8$\times$ \\
\bottomrule
\end{tabular}
\end{table}

\textbf{Evaluation.}
We evaluate each SAE on held-out activations excluded from every training buffer and use explained variance (EV) as the primary quality metric. The lower endpoint of a saturation interval is the smallest tested buffer whose EV is within $\epsilon=0.005$ of the maximum observed EV for that configuration. The upper endpoint is the next larger grid point. These two tested sizes describe the plateau range without estimating uncertainty or locating the threshold between measurements. We also evaluate downstream SAEBench metrics in Section~\ref{sec:saebench}.

\textbf{Training hyperparameters.}
Unless otherwise stated, all SAEs use the same optimizer, schedule, batch size, and activation normalization. The held-out EV set contains $10^6$ FineWeb tokens with no overlap with any training buffer. Full training hyperparameters are given in Appendix~\ref{app:hyperparams}.

\section{Results: The Diversity--Repetition Tradeoff}
\label{sec:results}

We first study how explained variance depends on the number of unique activations when the total SAE training budget is held fixed. For each total budget $N_{\mathrm{total}}$, we vary the unique activation count $N_{\mathrm{unique}}$ over a logarithmic grid and train for
\begin{equation}
    E
    =
    \frac{N_{\mathrm{total}}}{N_{\mathrm{unique}}}
\end{equation}
epochs. Thus, all points within a curve use the same number of total SAE optimization tokens, but differ in how much activation diversity is available.

\subsection{Unique Activations Improve EV Under a Fixed Token Budget}
\label{sec:unique_improves_ev}

At a fixed total token budget, reconstruction improves substantially as the number of unique activations increases from the smallest buffers (Figure~\ref{fig:ev_vs_unique}). Training for many epochs over a very small buffer therefore gives worse reconstruction than spending the same budget on fewer epochs over a larger buffer.

For the budget of 128M tokens, the effect is large across all dictionary sizes. For example, increasing the unique activation count from 1M to the full 128M budget improves EV from $0.5801$ to $0.6393$ for the 16k dictionary, and from $0.5870$ to $0.6610$ for the 32k dictionary. The 64k dictionary shows the same pattern, improving from $0.5860$ at 1M unique activations to $0.6648$ at 128M unique activations.

The curves at 1.28B tokens have higher absolute EV, but small buffers still limit reconstruction. Even at this larger budget, additional replay cannot fully compensate for insufficient activation coverage.

\begin{figure}[t]
    \centering
    \includegraphics[width=0.60\linewidth]{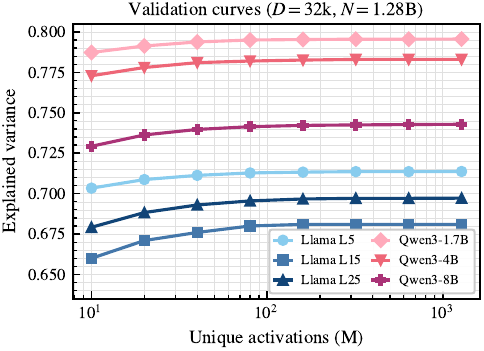}
    \caption{
    Validation across layers and models ($D=32\mathrm{k}$, $N=1.28\mathrm{B}$). Llama-3.1 8B layer sweeps use BatchTopK SAEs, and sweeps across Qwen3 model sizes use TopK SAEs. All curves show the same saturation pattern as diversity increases despite changes in layer, model size, and SAE activation function.
    }
    \label{fig:qwen_topk_validation}
\end{figure}

\subsection{Explained Variance Saturates in Unique Activation Count}
\label{sec:ev_saturates}

Across all dictionary sizes, EV rises rapidly as small buffers grow and then reaches a plateau. Once the buffer reaches this plateau, additional unique activations provide negligible reconstruction benefit. Training can therefore reuse a sufficiently diverse buffer without storing a fresh activation for every optimization token.

We define saturation as reaching EV within $\epsilon=0.005$ of the maximum observed EV on the corresponding curve. Because the tested buffer sizes form a logarithmic grid, we report the saturation threshold as an interval between adjacent grid points: the smallest observed buffer size on the plateau and the next larger grid point.

\begin{table}[t]
\centering
\caption{
Transfer across activation functions at $D=32\mathrm{k}$, $N_{\mathrm{total}}=640\mathrm{M}$, Llama-3.1 8B layer 15. BatchTopK and TopK are matched at $k=32$. JumpReLU converges to $L_0 \approx 100$ and is therefore less sparse. All three variants saturate at 40M unique activations, showing that the diversity--repetition tradeoff is not specific to BatchTopK.
}
\label{tab:activation_transfer}
\begin{tabular}{lrrrr}
\toprule
Activation & Sparsity & Plateau EV & $u_{\mathrm{sat}}$ & Storage reduction \\
\midrule
BatchTopK   & average $k=32$   & $0.6803$ & 40M  & $16\times$ \\
TopK        & $k=32$           & $0.6807$ & 40M  & $16\times$ \\
JumpReLU    & $L_0 \approx 100$ & $0.7634$ & 40M  & $16\times$ \\
\bottomrule
\end{tabular}
\end{table}

Two patterns emerge from Table~\ref{tab:saturation_intervals}. First, at the largest budget the saturation interval scales approximately linearly with dictionary size: 20M--40M for 16k, 40M--80M for 32k, 80M--160M for 64k, and 160M--320M for 128k. Second, once the total budget comfortably exceeds the saturation interval for that dictionary, the interval depends only weakly on $N_{\mathrm{total}}$. For example, the 16k, 32k, and 64k dictionaries have the same saturation intervals at 640M and 1.28B. The implied storage reduction therefore grows with total budget, reaching $\mathbf{64\times}$ for the 16k dictionary at 1.28B tokens.

\subsection{Larger Dictionaries Reach Higher Plateaus but Require More Diversity}
\label{sec:dict_size_effect}

Larger dictionaries reach higher EV but require more unique activations to saturate. At a budget of 1.28B tokens, doubling $D$ raises the best observed EV by roughly 2--3 EV points and moves the saturation interval upward by roughly $2\times$ (Table~\ref{tab:saturation_intervals}). The gain is smaller at lower budgets: the 128k dictionary reaches $0.711$ EV at 640M tokens and $0.720$ at 1.28B. This suggests that $0.711$ reflects incomplete training rather than the dictionary's asymptotic performance. Total token count alone cannot distinguish dictionary capacity from the diversity needed to train it. Since $E=N_{\mathrm{total}}/U$, the smallest buffers in Figure~\ref{fig:ev_vs_unique} also receive the most replay. Those extra passes do not make up for the missing diversity.

\subsection{Validation across layers and models}
\label{sec:cross_arch_results}

We test other layers and models at $D=32\mathrm{k}$ and $N_{\mathrm{total}}=1.28\mathrm{B}$, using the same fractions of unique activations as the main grid. These sweeps test transfer of the saturation pattern at one width and budget. They do not repeat the full scaling grid.

\textbf{Layer generalization.}
We isolate layer effects by training BatchTopK SAEs on Llama-3.1 8B residual stream activations from layers 5, 15, and 25. Each layer has a different plateau EV: $0.713$ at layer 5, $0.681$ at layer 15, and $0.697$ at layer 25. Despite these differences in representation structure, EV improves with activation diversity and then plateaus at all three layers.

\textbf{Generalization across model sizes.}
We train SAEs on Qwen3-1.7B, Qwen3-4B, and Qwen3-8B activations from comparable layers near the middle of each model. These runs use standard TopK to reduce the cost of the validation sweep. Both the base model and SAE activation function therefore differ from the main grid, so this comparison tests their combined effect. Plateau EV ranges from $0.743$ for Qwen3-8B to $0.796$ for Qwen3-1.7B. All three model sizes show saturation as activation diversity increases.

Figure~\ref{fig:qwen_topk_validation} shows all six curves: early, middle, and late Llama layers with BatchTopK, and three Qwen model sizes with TopK.

\subsection{Transfer across activation functions}
\label{sec:activation_transfer}

We compare activation functions on Llama-3.1 8B layer 15 at $D=32\mathrm{k}$ and $N_{\mathrm{total}}=640\mathrm{M}$. BatchTopK, standard TopK~\citep{gao2024scaling}, and JumpReLU~\citep{rajamanoharan2024jumping} use the same grid of buffer sizes, $u \in \{5,10,20,40,80,160,320,640\}$M. BatchTopK targets $k=32$ on average, while TopK uses $k=32$ active features per token. JumpReLU learns a threshold and converges to an average $L_0 \approx 100$. Its lower sparsity prevents a direct comparison of absolute EV with the two TopK variants.

\textbf{Saturation pattern.}
All three activation functions improve rapidly with the number of unique activations and reach saturation at 40M unique activations (Table~\ref{tab:activation_transfer}). Their different gates and plateau EV values leave the saturation pattern unchanged at $D=32$k. The tradeoff between diversity and repetition therefore extends beyond BatchTopK in this setting.

\textbf{Plateau levels.}
At matched $k=32$, TopK reaches $0.681$ EV and BatchTopK reaches $0.680$. JumpReLU reaches $0.763$ with many more active latents per token ($L_0 \approx 100$). That difference in sparsity prevents attributing the higher EV to the gate alone. The experiment establishes transfer of the saturation pattern without ranking the gates at a common sparsity.

\begin{figure}[t]
    \centering
    \includegraphics[width=\linewidth]{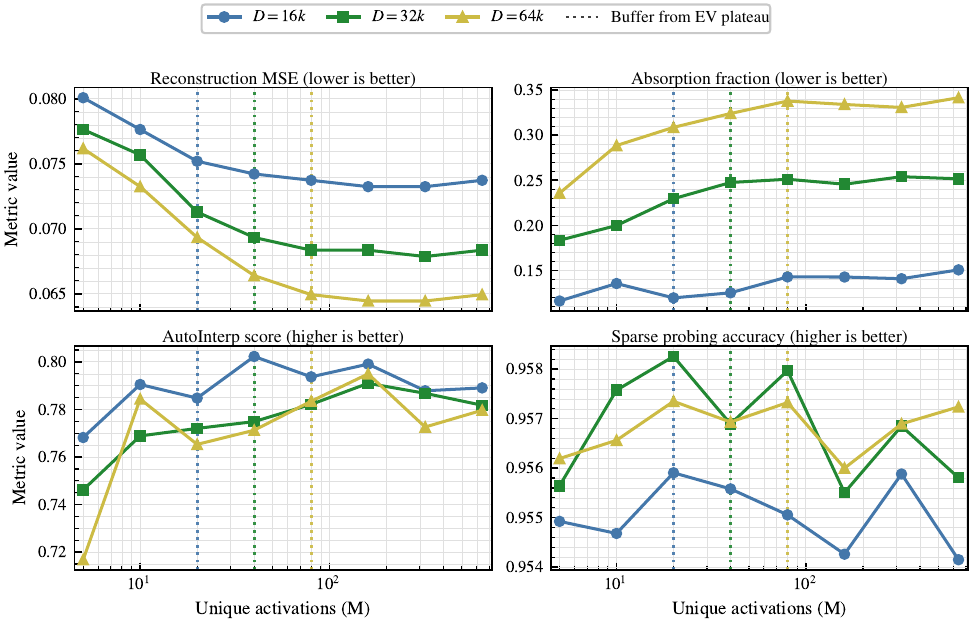}
    \caption{
    SAEBench metrics versus the number of unique activations for the BatchTopK sweep at layer 15 with a budget of 640M tokens. Dotted vertical lines mark the buffer chosen from the EV plateau for each dictionary. Reconstruction MSE follows the EV plateau, and AutoInterp remains high beyond this buffer scale. Feature absorption measures disentanglement. Sparse probing changes little with buffer size. None of the evaluated downstream metrics drops sharply at the recommended buffer size. Lower MSE and absorption fraction are better. Higher AutoInterp and sparse probing accuracy are better.
    }
    \label{fig:saebench}
\end{figure}

\subsection{Downstream Interpretability Tracks EV Saturation}
\label{sec:saebench}

To check the downstream effects of choosing a buffer by EV, we evaluate 24 checkpoints from the BatchTopK sweep at layer 15 with a budget of 640M tokens on SAEBench. There are eight buffer sizes for each of $D \in \{16\mathrm{k},32\mathrm{k},64\mathrm{k}\}$. Dotted vertical lines in Figure~\ref{fig:saebench} mark the buffer chosen using EV for each dictionary, allowing comparison with the reconstruction and interpretability metrics.

\textbf{Comparison to the EV buffer scale.}
Reconstruction MSE improves rapidly at low diversity and is near its best value at the buffer chosen using EV. EV rescales reconstruction error on a fixed evaluation set, so MSE checks consistency with the reconstruction result. Automated interpretability remains high across the sweep, with only modest further changes as the buffer grows beyond the EV recommendation. Table~\ref{tab:saebench} reports thresholds for each metric, including AutoInterp at 40M, 160M, and 160M for the three widths. Sparse probing accuracy is nearly flat. Absorption fraction, for which lower values are better, does not improve monotonically with diversity: the 32k and 64k curves rise. Reconstruction and disentanglement therefore require separate evaluation.

\textbf{Implications.}
AutoInterp and the other downstream metrics support using EV saturation to choose the buffer, even when their optima occur elsewhere on the grid. Both absolute and relative changes with buffers larger than the recommendation are small. At this scale, reconstruction quality is preserved, feature absorption and automated interpretability stay in the same narrow performance band, and sparse probing is largely unaffected.

\subsection{Math and code}
\label{sec:domain_transfer}
We change the activation source while keeping Llama-3.1 8B layer 15, BatchTopK, $D=32$k, $k=32$, and 640M total presentations fixed. Evaluation uses held-out activations from the same domain. FineMath first meets $\epsilon=0.005$ at 80M unique activations, relative to its run without replay. Stack-Edu Python meets the same tolerance at 40M (Table~\ref{tab:domain_transfer}). The corresponding storage reductions are $8\times$ and $16\times$. Saturation thus persists in both domains, although the tested counts cannot locate a continuous threshold or establish transfer of the FineWeb coefficients.
\begin{table}[t]
\centering
\caption{Additional domain sweeps: held-out EV at each tested buffer size, with 640M total presentations. The reference is each domain's 640M run without replay.}
\label{tab:domain_transfer}
\small
\begin{tabular}{lrrrrrr}
\toprule
Domain & 10M & 20M & 40M & 80M & 640M & $U_{0.005}$ / reduction \\
\midrule
FineMath-3plus & 0.6312 & 0.6523 & 0.6608 & 0.6649 & 0.6673 & 80M / $8\times$ \\
Stack-Edu Python & 0.7013 & 0.7081 & 0.7119 & 0.7143 & 0.7155 & 40M / $16\times$ \\
\bottomrule
\end{tabular}
\end{table}

\subsection{Tolerance sensitivity}
\label{sec:tolerance}
Table~\ref{tab:tolerance} reports the first qualifying buffers at half and twice the default EV tolerance. Tightening $\epsilon$ to $0.0025$ doubles the Python buffer and halves its storage savings. Loosening it to $0.010$ halves the buffer and doubles the savings. FineMath selects the same buffer under the stricter tolerance because of the spacing between tested counts. The 640M sweeps confirm the default recommendations and test sensitivity across widths (Appendix~\ref{app:recovered}). A buffer recommendation therefore depends on its reference EV, tested grid, and tolerance.
\begin{table}[t]
\centering
\caption{First qualifying tested buffer / storage reduction at three absolute EV tolerances, relative to the domain's 640M run.}
\label{tab:tolerance}
\begin{tabular}{lrrr}
\toprule
Domain & $\epsilon=0.0025$ & $\epsilon=0.005$ & $\epsilon=0.010$ \\
\midrule
FineMath-3plus & 80M / $8\times$ & 80M / $8\times$ & 40M / $16\times$ \\
Stack-Edu Python & 80M / $8\times$ & 40M / $16\times$ & 20M / $32\times$ \\
\bottomrule
\end{tabular}
\end{table}

\subsection{Effective Activation Count}
\label{sec:neff}

When a replay buffer contains $U$ unique activations and is used for $N_{\mathrm{total}}$ training tokens, each activation is replayed on average $N_{\mathrm{total}}/U$ times. We define the repetition count as
\begin{equation}
    R = \frac{N_{\mathrm{total}}}{U} - 1 .
    \label{eq:repetitions}
\end{equation}
Following~\citet{muennighoff2023scaling}, we convert $(U,N_{\mathrm{total}})$ into an effective activation count
\begin{equation}
    N_{\mathrm{eff}}(U,N_{\mathrm{total}},R^*)
    =
    U \left[1 + R^*\left(1-e^{-R/R^*}\right)\right],
    \label{eq:neff}
\end{equation}
where $R^*>0$ is the fitted characteristic repetition scale. We obtain this form by assuming that the marginal contribution of replay decays as $e^{-r/R^*}$ and integrating from $r=0$ to $R$. The effective count is a phenomenological discount on repeated data and does not measure the number of features. When $R \ll R^*$, replay is nearly as useful as fresh data and $N_{\mathrm{eff}}\approx N_{\mathrm{total}}$. When $R \gg R^*$, further replay adds little, and $N_{\mathrm{eff}}$ is limited to a constant multiple of $U$.

\subsection{Joint Scaling Law}
\label{sec:joint_law}

We model explained variance as
\begin{equation}
    \mathrm{EV}(D,N_{\mathrm{eff}})
    =
    E_\infty
    -
    A D^{-\alpha}
    -
    B N_{\mathrm{eff}}^{-\beta}.
    \label{eq:joint_law}
\end{equation}
The extrapolated ceiling $E_\infty$ depends on the model, layer, normalization, objective, and sparsity. Its fitted value of $0.760$ exceeds the best EV observed on the main grid, $0.7197$. The first penalty describes finite dictionary capacity and decreases as the dictionary grows. The second describes limited effective data. At a fixed training budget, larger buffers require fewer repetitions and reduce this penalty.

For larger dictionaries, activation diversity appears to become limiting after fewer replay passes. We model this with a repetition scale $R^*$ that depends on dictionary size:
\begin{equation}
    R^*(D) = cD^{-\gamma}.
    \label{eq:rstar_D}
\end{equation}
The coefficient $c$ sets the repetition scale, and $\gamma$ determines how it changes with width. With the fitted $\gamma=0.526$, doubling the width multiplies $R^*$ by $2^{-0.526}\approx0.69$. Larger dictionaries therefore tolerate fewer replay passes before diversity becomes limiting.

\subsection{Fitting and Model Comparison}
\label{sec:fitting}

We fit to the 96 observations in the main grid: three total budgets, four dictionary sizes, and eight fractions of unique activations for Llama-3.1 8B layer 15 BatchTopK SAEs. We compare the models with constant $R^*$ and with $R^*(D)$ using the Akaike Information Criterion (AIC). Appendix~\ref{app:fitting} describes the fitting procedure.

Allowing the repetition scale to depend on dictionary size reduces the fit error to $\mathrm{RMSE}=4.8\times10^{-3}$ EV and improves AIC. This comparison supports a dependence between dictionary size and the benefit of activation diversity. Appendix~\ref{app:fitting} gives the full fitted parameters.

\begin{figure}[t]
    \centering
    \includegraphics[width=\linewidth]{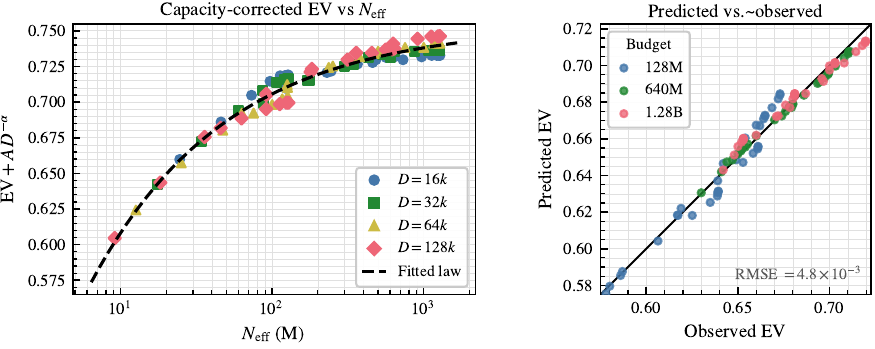}
    \caption{
    Capacity-corrected EV ($\mathrm{EV} + A D^{-\alpha}$) versus effective activation count $N_{\mathrm{eff}}$ under the extended $R^*(D)$ model. Data points from all four dictionary sizes and three total budgets collapse onto a single power-law curve, supporting the functional form within the fitted regime.
    }
    \label{fig:saturation_scaling}
\end{figure}

After removing the fitted capacity term, the 96 data points follow a common power-law curve: Figure~\ref{fig:saturation_scaling} (left) plots $\mathrm{EV} + A D^{-\alpha}$ against $N_{\mathrm{eff}}(D)$. The right panel compares predicted and observed EV, with $\mathrm{RMSE} = 4.8 \times 10^{-3}$.

\subsection{Validation}
\label{sec:params}

The common curve in Figure~\ref{fig:saturation_scaling} describes the fitted grid as a function of $N_{\mathrm{eff}}$. We examine the residuals within individual panels and test predictions for budgets, widths, layers, and models excluded from the fit.

\textbf{Panel residuals and buffer decisions.}
Fit errors are largest at the smallest budget and at the extreme widths of the largest budget (Table~\ref{tab:fit_residuals}). Undertraining and the smooth law's approximation to steep curves at low diversity may contribute to these residuals. Fitted and observed plateau intervals coincide in six of the twelve configurations. In the remaining six, the fitted interval is one grid step larger, leading to a larger buffer recommendation. This conservative direction of error on the measured grid need not hold in new settings.

\begin{table}[t]
\centering
\caption{RMSE for each panel used in the fit ($\times10^{-3}$ EV), with eight buffer sizes per cell. Residuals concentrate at the smallest budget and extreme widths at the largest budget.}
\label{tab:fit_residuals}
\begin{tabular}{lrrrr}
\toprule
Budget & 16k & 32k & 64k & 128k \\
\midrule
128M & 7.7 & 5.0 & 6.1 & 8.1 \\
640M & 1.8 & 1.6 & 2.9 & 2.4 \\
1.28B & 5.9 & 2.5 & 1.6 & 5.2 \\
\bottomrule
\end{tabular}
\end{table}

\textbf{Held-out training budget.}
Fitting the law on the two extreme budgets, 128M and 1.28B tokens, gives predictions for the intermediate 640M budget with test RMSE $3.1\times10^{-3}$ EV. Predicted saturation intervals agree with the observed intervals to within one grid step.

\textbf{Held-out dictionary size.}
We fit on the three smaller dictionaries and predict EV for the held-out 128k dictionary. Extrapolation succeeds at 1.28B tokens, where that dictionary is sufficiently trained, but becomes less accurate at smaller budgets. In the undertrained regime, its best observed EV has not reached the asymptotic plateau. A fit to smaller dictionaries that have received sufficient training does not capture this regime.

\textbf{Shape transfer.}
For the curves from other layers and models, we keep the nonlinear shape fixed and refit the level parameters. The validation fits have RMSE at most $0.9\times10^{-3}$ EV, supporting reuse of the curve shape after this recalibration. Because all these checks use one dictionary width, they cannot establish transfer of width exponents or unchanged coefficients to new domains.

\section{Practical Implications: Activation Buffer Sizing}
\label{sec:practical}

The saturation intervals determine how much activation storage a training run needs. A buffer that reaches the quality plateau can be replayed for the remaining budget, avoiding the storage of one fresh activation per training token.

For dictionary size $D$ and total SAE training budget $N_{\mathrm{total}}$, choose $U_{\mathrm{buf}}$ from the corresponding saturation interval. Table~\ref{tab:storage_recipe} uses each interval's lower endpoint: the smallest tested buffer whose EV is within $\epsilon=0.005$ of the best value on the curve. The number of replay epochs is
\begin{equation}
    E = \frac{N_{\mathrm{total}}}{U_{\mathrm{buf}}},
    \label{eq:epochs}
\end{equation}
and the storage reduction relative to storing all activations is
\begin{equation}
    \mathrm{StorageReduction}
    =
    \frac{N_{\mathrm{total}}}{U_{\mathrm{buf}}}.
    \label{eq:storage_reduction}
\end{equation}

Every configuration in our grid permits at least an $8\times$ reduction in activation storage. Smaller dictionaries at the largest budget reach $64\times$ (Appendix~\ref{app:storage_recipe}), reducing several terabytes of activations to a few hundred gigabytes. The 128k dictionary at 1.28B tokens still needs a 160M buffer (1.19 TiB), an $8\times$ reduction from the 9.54 TiB full corpus. A Llama-3.1 8B residual stream in bf16 takes 8\,KiB per activation. At $N_{\mathrm{total}}=1.28\mathrm{B}$ and $D=16\mathrm{k}$, the full corpus requires $9.54\,\mathrm{TiB}$ of storage, compared with $152.6\,\mathrm{GiB}$ for the recommended buffer.

Storage savings grow with total budget once the dictionary has enough activation diversity to saturate. Smaller dictionaries reach this point early, so further increases in $N_{\mathrm{total}}$ mainly add replay. Larger dictionaries at small budgets can remain limited by capacity or optimization. Their savings grow more slowly until the budget is sufficient to reach the full plateau.

\textbf{Training recipe.}
For a chosen dictionary size and total budget, estimate the saturation interval with a pilot sweep or the fitted law. Set $U_{\mathrm{buf}}$ to the smallest buffer on the plateau and train for $N_{\mathrm{total}}/U_{\mathrm{buf}}$ replay epochs.

\Needspace{8\baselineskip}
\textbf{Stored buffers and fresh streaming.}
When storage permits caching, generating $U_{\mathrm{buf}}$ once and replaying it reduces the work of generating activations relative to generating fresh activations on the fly. SAE optimization still uses $N_{\mathrm{total}}$ presentations. For generation and SAE costs per token $C_{\mathrm{gen}}$ and $C_{\mathrm{SAE}}$, the idealized compute ratio is
\begin{equation}
\frac{N_{\mathrm{total}}(C_{\mathrm{gen}}+C_{\mathrm{SAE}})}
{U_{\mathrm{buf}}C_{\mathrm{gen}}+N_{\mathrm{total}}C_{\mathrm{SAE}}}.
\label{eq:compute_ratio}
\end{equation}
This ratio approaches the storage ratio when activation generation dominates the cost. We do not measure end-to-end runtime. Recomputing activations on every replay pass preserves the diversity schedule but requires the same number of generation forwards. Fresh streaming remains useful when storage is constrained or the distribution changes. The storage ratio equals the number of passes: both are $N_{\mathrm{total}}/U_{\mathrm{buf}}$.

\section{Conclusion}
At a fixed SAE training budget, activation diversity limits reconstruction until the buffer reaches a quality plateau. Replay beyond that point preserves reconstruction and downstream interpretability performance. We observe this behavior in the 96-run scaling grid and validate it on held-out settings, other layers and models, and different SAE gates. The experiments on mathematical text and Python code extend the domain coverage. Choosing the smallest tested buffer within the EV tolerance yields $8\text{--}64\times$ savings in activation storage.

\textbf{Limitations and future work.}
The full grid uses FineWeb and Llama-3.1 8B at fixed sparsity. The domain sweeps test qualitative transfer, while the layer and model sweeps test curve shape after recalibration. These checks leave the generality of the coefficients unresolved. Extrapolation is less accurate for undertrained wider dictionaries, and stricter EV tolerances can require larger buffers. Gate comparisons cover one width. SAEBench covers the BatchTopK sweep at layer 15. We have not tested objectives across tokens, broader modalities, or end-to-end runtime. Future work could examine adaptive buffer refreshing and fits that account for sparsity.

\clearpage
\bibliographystyle{plainnat}
\bibliography{references}

\appendix

\newpage

\section{Additional Experimental Details}
\label{app:hyperparams}

\paragraph{Training hyperparameters.}
Unless otherwise stated, SAEs are trained with AdamW ($\beta_1=0.9$, $\beta_2=0.99$, weight decay $0.05$), peak learning rate $3\times10^{-4}$, cosine decay to $10^{-6}$, and a $10\%$ linear warmup. Batch size is 4096 activations and gradients are clipped to unit norm. For TopK variants, $k$ is linearly annealed from $128$ to the target value over the first $10\%$ of training. BatchTopK runs use an auxiliary penalty for dead latents of $1/32$ and straight-through estimator bandwidth $0.001$. We normalize all activations to unit norm before passing them to the SAE.

\section{Scaling Law Fitting Details}
\label{app:fitting}

With the nonlinear parameters fixed, Eq.~\ref{eq:joint_law} is linear in $(E_\infty,A,B)$. At each candidate set of nonlinear parameters, we solve for the linear parameters by ordinary least squares, then refine the best solution with local coordinate descent. The nonlinear parameters are $(\alpha,\beta,R^*)$ for the baseline with constant $R^*$ and $(\alpha,\beta,c,\gamma)$ for the model whose repetition scale varies with dictionary size. We fit both to the 96-point main grid and compare them using the Akaike Information Criterion (AIC).

\begin{table}[htbp]
\centering
\caption{
Fitted parameters for the model with constant $R^*$ and the extended $R^*(D)$ model, fit on all 96 data points from the main grid. The extended model is decisively preferred by AIC.
}
\label{tab:fitted_params}
\begin{tabular}{lrrrrrrrr}
\toprule
Model & $E_\infty$ & $A$ & $\alpha$ & $B$ & $\beta$ & $R^*$ / $c$ & $\gamma$ & $\Delta$AIC \\
\midrule
Const.\ $R^*$     & $0.738$ & $0.387$ & $0.688$ & $0.432$ & $0.466$ & $13.7$  & ---     & $0$ \\
Extended $R^*(D)$ & $0.760$ & $0.328$ & $0.486$ & $0.440$ & $0.477$ & $103.3$ & $0.526$ & $-77.3$ \\
\bottomrule
\end{tabular}
\end{table}

\section{SAEBench Threshold Table}
\label{app:saebench_thresholds}

For each SAEBench metric, Table~\ref{tab:saebench} gives the smallest buffer whose score is within 5\% of the metric's observed range from the best value on that curve. These thresholds describe the individual metrics. The main rule for sizing the buffer uses the saturation intervals determined by EV.

\begin{table}[htbp]
\centering
\caption{
SAEBench saturation thresholds at $N_{\mathrm{total}}=640\mathrm{M}$, Llama-3.1 8B layer 15, BatchTopK. The EV row is reproduced from Table~\ref{tab:saturation_intervals}.
}
\label{tab:saebench}
\begin{tabular}{lrrr}
\toprule
Quantity                            & $D=16$k & $D=32$k & $D=64$k \\
\midrule
Explained variance (main)           & 20M  & 40M  & 80M  \\
Reconstruction MSE                  & 80M  & 80M  & 80M  \\
Feature absorption fraction         & 80M  & 40M  & 80M  \\
Automated interpretability score    & 40M  & 160M & 160M \\
SAE sparse probing accuracy         & 20M  & 20M  & 20M  \\
\bottomrule
\end{tabular}
\end{table}

\clearpage
\section{Storage Reduction Recommendations}
\label{app:storage_recipe}

\begin{table}[ht!]
\centering
\caption{
Recommended activation buffer sizes and implied storage reductions ($\epsilon = 0.005$). Epochs $= N_{\mathrm{total}} / u_{\mathrm{sat}}$.
}
\label{tab:storage_recipe}
\begin{tabular}{lrrrr}
\toprule
Budget & Dict.\ size & $u_{\mathrm{sat}}$ & Epochs & Storage reduction \\
\midrule
128M  & 16k  & 8M   & 16 & $16\times$ \\
128M  & 32k  & 16M  & 8  & $8\times$ \\
128M  & 64k  & 16M  & 8  & $8\times$ \\
128M  & 128k & 16M  & 8  & $8\times$ \\
640M  & 16k  & 20M  & 32 & $32\times$ \\
640M  & 32k  & 40M  & 16 & $16\times$ \\
640M  & 64k  & 80M  & 8  & $8\times$ \\
640M  & 128k & 80M  & 8  & $8\times$ \\
1.28B & 16k  & 20M  & 64 & $64\times$ \\
1.28B & 32k  & 40M  & 32 & $32\times$ \\
1.28B & 64k  & 80M  & 16 & $16\times$ \\
1.28B & 128k & 160M & 8  & $8\times$ \\
\bottomrule
\end{tabular}
\end{table}

\section{Additional Results on Threshold Sensitivity}
\label{app:recovered}
The domain sweeps use \texttt{HuggingFaceTB/finemath} with configuration \texttt{finemath-3plus}, and \texttt{HuggingFaceTB/stack-edu} with configuration \texttt{Python}. The datasets are described by \citet{allal2025smollm2}. The sweeps use Llama-3.1 8B layer 15, BatchTopK, $D=32$k, $k=32$, and 640M total presentations. Held-out evaluation uses activations from the same domain, and the tolerance reference is that domain's 640M run without replay.

The comparison across dictionary widths uses a budget of 640M total presentations and the best final EV on each sweep as its reference. Stricter tolerances can require larger buffers and reduce the storage savings. The default recommendations agree with those in the main saturation table.

\begin{table}[htbp]
\centering
\caption{Buffer recommendations at 640M total presentations for three EV tolerances. The reference is the best final EV on each sweep over eight buffer sizes. Each cell gives the smallest qualifying buffer and the resulting storage reduction.}
\label{tab:recovered_tolerance}
\begin{tabular}{lrrr}
\toprule
Width & $\epsilon=0.0025$ & $\epsilon=0.005$ & $\epsilon=0.010$ \\
\midrule
16k & 40M / $16\times$ & 20M / $32\times$ & 20M / $32\times$ \\
32k & 80M / $8\times$ & 40M / $16\times$ & 20M / $32\times$ \\
64k & 80M / $8\times$ & 80M / $8\times$ & 40M / $16\times$ \\
128k & 160M / $4\times$ & 80M / $8\times$ & 40M / $16\times$ \\
\bottomrule
\end{tabular}
\end{table}

\clearpage
\input{checklist.tex}

\end{document}

%% file: figures/replay_pipeline.tex
\begin{tikzpicture}[
box/.style={draw=black!55,rounded corners=2pt,align=center,text width=5.2cm,minimum height=0.75cm,font=\small,fill=black!3},
flow/.style={-{Stealth[length=2mm]},thick},node distance=0.38cm]
\node[font=\small\bfseries] (lt) at (0,0) {Fresh activations};
\node[font=\small\bfseries] (rt) at (6.4,0) {Buffer replay};
\node[box,below=0.2cm of lt] (fresh) {Generate $N_{\mathrm{total}}$ activations};
\node[box,below=of fresh] (full) {Store $N_{\mathrm{total}}$ vectors\\Footprint: $sN_{\mathrm{total}}$ bytes};
\node[box,below=of full] (once) {One pass through the corpus\\$N_{\mathrm{total}}$ SAE presentations};
\node[box,below=0.2cm of rt] (sample) {Generate $U_{\mathrm{buf}}$ activations};
\node[box,below=of sample] (buffer) {Store $U_{\mathrm{buf}}$ vectors\\Footprint: $sU_{\mathrm{buf}}$ bytes};
\node[box,below=of buffer] (repeat) {$E=N_{\mathrm{total}}/U_{\mathrm{buf}}$ passes\\$N_{\mathrm{total}}$ SAE presentations};
\draw[flow] (fresh)--(full);\draw[flow] (full)--(once);
\draw[flow] (sample)--(buffer);\draw[flow] (buffer)--(repeat);
\draw[flow] (repeat.east)--++(0.25,0)|-(buffer.east);
\node[align=center,font=\small,below=0.25cm of once,xshift=3.2cm]
{Same SAE budget. Storage ratio $N_{\mathrm{total}}/U_{\mathrm{buf}}=E$};
\end{tikzpicture}

%% file: checklist.tex
\section*{NeurIPS Paper Checklist}

\begin{enumerate}

\Needspace{9\baselineskip}
\item {\bf Claims}
    \item[] Question: Do the main claims made in the abstract and introduction accurately reflect the paper's contributions and scope?
    \item[] Answer: \answerYes{}
    \item[] Justification: The abstract and introduction describe the observed saturation, the settings tested, the fitted law, and the storage savings. The limitations paragraph states where these findings may not generalize.

\Needspace{9\baselineskip}
\item {\bf Limitations}
    \item[] Question: Does the paper discuss the limitations of the work performed by the authors?
    \item[] Answer: \answerYes{}
    \item[] Justification: The paper includes a limitations paragraph discussing the main grid, targeted domain transfer, fixed sparsity, prediction limits, tolerance sensitivity, and limited downstream scope.

\Needspace{9\baselineskip}
\item {\bf Theory assumptions and proofs}
    \item[] Question: For each theoretical result, does the paper provide the full set of assumptions and a complete (and correct) proof?
    \item[] Answer: \answerNA{}
    \item[] Justification: The paper proposes and fits a phenomenological scaling law but does not present formal theoretical results, theorems, or proofs.

\Needspace{9\baselineskip}
\item {\bf Experimental result reproducibility}
    \item[] Question: Does the paper fully disclose all the information needed to reproduce the main experimental results of the paper to the extent that it affects the main claims and/or conclusions of the paper (regardless of whether the code and data are provided or not)?
    \item[] Answer: \answerNo{}
    \item[] Justification: The setup and appendices describe the experimental design, hyperparameters, held-out FineWeb size, domain sweeps, and threshold sensitivity. Complete training, evaluation, and fitting code and full split and run metadata are not provided. Appendix~\ref{app:recovered} reports additional threshold sensitivity results.

\Needspace{9\baselineskip}
\item {\bf Open access to data and code}
    \item[] Question: Does the paper provide open access to the data and code, with sufficient instructions to faithfully reproduce the main experimental results, as described in supplemental material?
    \item[] Answer: \answerNo{}
    \item[] Justification: The supplemental files include reported results, EV measurements, and scripts for checking summary calculations. A complete implementation for training, evaluation, and fitting is not included.

\Needspace{9\baselineskip}
\item {\bf Experimental setting/details}
    \item[] Question: Does the paper specify all the training and test details (e.g., data splits, hyperparameters, how they were chosen, type of optimizer) necessary to understand the results?
    \item[] Answer: \answerNo{}
    \item[] Justification: The models, datasets, layer-15 setting, budgets, grid, optimizer, normalization, and the FineWeb evaluation set of one million tokens are described. The sizes of domain evaluation sets, detailed split construction, model/dataset revisions, and evaluator settings are not fully specified. Appendix~\ref{app:recovered} reports additional threshold sensitivity results.

\Needspace{9\baselineskip}
\item {\bf Experiment statistical significance}
    \item[] Question: Does the paper report error bars suitably and correctly defined or other appropriate information about the statistical significance of the experiments?
    \item[] Answer: \answerNo{}
    \item[] Justification: The main grid uses single training runs. We do not report error bars or confidence intervals across random seeds.

\Needspace{9\baselineskip}
\item {\bf Experiments compute resources}
    \item[] Question: For each experiment, does the paper provide sufficient information on the computer resources (type of compute workers, memory, time of execution) needed to reproduce the experiments?
    \item[] Answer: \answerNo{}
    \item[] Justification: The paper describes the training scale and requirements for activation storage, but does not provide a complete accounting of GPU type, memory, wall-clock time, or total compute for each experiment.

\Needspace{9\baselineskip}
\item {\bf Code of ethics}
    \item[] Question: Does the research conducted in the paper conform, in every respect, with the NeurIPS Code of Ethics \url{https://neurips.cc/public/EthicsGuidelines}?
    \item[] Answer: \answerYes{}
    \item[] Justification: The work is an empirical study of SAE training efficiency using publicly available models and text/code datasets, and we are not aware of any deviation from the NeurIPS Code of Ethics.

\Needspace{9\baselineskip}
\item {\bf Broader impacts}
    \item[] Question: Does the paper discuss both potential positive societal impacts and negative societal impacts of the work performed?
    \item[] Answer: \answerNA{}
    \item[] Justification: The work is foundational research on reducing costs of activation storage for sparse autoencoder training and does not introduce a deployed system or direct application pathway. Its main expected impact is improved accessibility of interpretability experiments through lower storage costs.

\Needspace{9\baselineskip}
\item {\bf Safeguards}
    \item[] Question: Does the paper describe safeguards that have been put in place for responsible release of data or models that have a high risk for misuse (e.g., pre-trained language models, image generators, or scraped datasets)?
    \item[] Answer: \answerNA{}
    \item[] Justification: The paper does not release a new high-risk pretrained model, image generator, or scraped dataset requiring special safeguards.

\Needspace{9\baselineskip}
\item {\bf Licenses for existing assets}
    \item[] Question: Are the creators or original owners of assets (e.g., code, data, models), used in the paper, properly credited and are the license and terms of use explicitly mentioned and properly respected?
    \item[] Answer: \answerNo{}
    \item[] Justification: The paper cites the main existing assets used, including the named base models, FineWeb, FineMath, Stack-Edu, SAEBench, and prior SAE methods, but it does not explicitly enumerate all asset licenses and terms of use.

\Needspace{9\baselineskip}
\item {\bf New assets}
    \item[] Question: Are new assets introduced in the paper well documented and is the documentation provided alongside the assets?
    \item[] Answer: \answerNA{}
    \item[] Justification: The supplemental files document the reported results and scripts for summary calculations and figure annotations. They contain no new training dataset, benchmark, or model.

\Needspace{9\baselineskip}
\item {\bf Crowdsourcing and research with human subjects}
    \item[] Question: For crowdsourcing experiments and research with human subjects, does the paper include the full text of instructions given to participants and screenshots, if applicable, as well as details about compensation (if any)? 
    \item[] Answer: \answerNA{}
    \item[] Justification: The paper does not involve crowdsourcing or research with human subjects.

\Needspace{9\baselineskip}
\item {\bf Institutional review board (IRB) approvals or equivalent for research with human subjects}
    \item[] Question: Does the paper describe potential risks incurred by study participants, whether such risks were disclosed to the subjects, and whether Institutional Review Board (IRB) approvals (or an equivalent approval/review based on the requirements of your country or institution) were obtained?
    \item[] Answer: \answerNA{}
    \item[] Justification: The paper does not involve crowdsourcing or research with human subjects, so IRB approval is not applicable.

\Needspace{9\baselineskip}
\item {\bf Declaration of LLM usage}
    \item[] Question: Does the paper describe the usage of LLMs if it is an important, original, or non-standard component of the core methods in this research? Note that if the LLM is used only for writing, editing, or formatting purposes and does \emph{not} impact the core methodology, scientific rigor, or originality of the research, declaration is not required.
    \item[] Answer: \answerNA{}
    \item[] Justification: The evaluation of automated interpretability uses an existing benchmark method that relies on an LLM. It does not introduce an original LLM component.

\end{enumerate}

%% file: references.bib
@inproceedings{cunningham2023sparse,
  title={Sparse autoencoders find highly interpretable features in language models},
  author={Cunningham, Hoagy and Ewart, Aidan and Riggs, Logan and Huben, Robert and Sharkey, Lee},
  year={2024},
  url={https://proceedings.iclr.cc/paper_files/paper/2024/file/1fa1ab11f4bd5f94b2ec20e794dbfa3b-Paper-Conference.pdf},
  booktitle={International Conference on Learning Representations}
}

@article{templeton2024scaling,
  title={Scaling Monosemanticity: Extracting Interpretable Features from {Claude 3 Sonnet}},
  author={Templeton, Adly and Conerly, Tom and Marcus, Jonathan and Lindsey, Jack and Bricken, Trenton and Chen, Brian and Pearce, Adam and Citro, Craig and Ameisen, Emmanuel and Jones, Andy and Cunningham, Hoagy and Turner, Nicholas L. and McDougall, Callum and MacDiarmid, Monte and Tamkin, Alex and Durmus, Esin and Hume, Tristan and Mosconi, Francesco and Freeman, C. Daniel and Sumers, Theodore R. and Rees, Edward and Batson, Joshua and Jermyn, Adam and Carter, Shan and Olah, Chris and Henighan, Tom},
  year={2024},
  url={https://transformer-circuits.pub/2024/scaling-monosemanticity/index.html},
  journal={Transformer Circuits Thread}
}

@misc{bills2023language,
  title={Language models can explain neurons in language models},
  author={Bills, Steven and Cammarata, Nick and Mossing, Dan and Tillman, Henk and Gao, Leo and Goh, Gabriel and Sutskever, Ilya and Leike, Jan and Wu, Jeff and Saunders, William},
  year={2023},
  url={https://openaipublic.blob.core.windows.net/neuron-explainer/paper/index.html},
  howpublished={OpenAI}
}

@inproceedings{paulo2024automatically,
  title={Automatically Interpreting Millions of Features in Large Language Models},
  author={Paulo, Gon{\c{c}}alo Santos and Mallen, Alex Troy and Juang, Caden and Belrose, Nora},
  year={2025},
  url={https://proceedings.mlr.press/v267/paulo25a.html},
  booktitle={Proceedings of the 42nd International Conference on Machine Learning},
  volume={267},
  pages={48393--48421},
  publisher={PMLR}
}

@article{lieberum2024gemma,
  title={{Gemma Scope}: Open Sparse Autoencoders Everywhere All At Once on {Gemma 2}},
  author={Lieberum, Tom and Rajamanoharan, Senthooran and Conmy, Arthur and Smith, Lewis and Sonnerat, Nicolas and Varma, Vikrant and Kram{\'a}r, J{\'a}nos and Dragan, Anca and Shah, Rohin and Nanda, Neel},
  journal={arXiv preprint arXiv:2408.05147},
  year={2024},
  url={https://arxiv.org/abs/2408.05147}
}

@inproceedings{gao2024scaling,
  title={Scaling and Evaluating Sparse Autoencoders},
  author={Gao, Leo and {Dupr\'e la Tour}, Tom and Tillman, Henk and Goh, Gabriel and Troll, Rajan and Radford, Alec and Sutskever, Ilya and Leike, Jan and Wu, Jeffrey},
  booktitle={International Conference on Learning Representations},
  year={2025},
  url={https://proceedings.iclr.cc/paper_files/paper/2025/file/42ef3308c230942d223c411adf182c88-Paper-Conference.pdf}
}

@inproceedings{karvonen2024saebench,
  title={{SAEBench}: A Comprehensive Benchmark for Sparse Autoencoders in Language Model Interpretability},
  author={Karvonen, Adam and Rager, Can and Lin, Johnny and Tigges, Curt and Bloom, Joseph Isaac and Chanin, David and Lau, Yeu-Tong and Farrell, Eoin and McDougall, Callum Stuart and Ayonrinde, Kola and Till, Demian and Wearden, Matthew and Conmy, Arthur and Marks, Samuel and Nanda, Neel},
  year={2025},
  url={https://proceedings.mlr.press/v267/karvonen25a.html},
  booktitle={Proceedings of the 42nd International Conference on Machine Learning},
  volume={267},
  pages={29223--29264},
  publisher={PMLR}
}

@inproceedings{muennighoff2023scaling,
  title={Scaling Data-Constrained Language Models},
  author={Muennighoff, Niklas and Rush, Alexander M. and Barak, Boaz and {Le Scao}, Teven and Piktus, Aleksandra and Tazi, Nouamane and Pyysalo, Sampo and Wolf, Thomas and Raffel, Colin},
  year={2023},
  url={https://proceedings.neurips.cc/paper_files/paper/2023/file/9d89448b63ce1e2e8dc7af72c984c196-Paper-Conference.pdf},
  booktitle={Advances in Neural Information Processing Systems},
  volume={36}
}

@inproceedings{hoffmann2022training,
  title={Training Compute-Optimal Large Language Models},
  author={Hoffmann, Jordan and Borgeaud, Sebastian and Mensch, Arthur and Buchatskaya, Elena and Cai, Trevor and Rutherford, Eliza and {de Las Casas}, Diego and Hendricks, Lisa Anne and Welbl, Johannes and Clark, Aidan and Hennigan, Tom and Noland, Eric and Millican, Katie and {van den Driessche}, George and Damoc, Bogdan and Guy, Aurelia and Osindero, Simon and Simonyan, Karen and Elsen, Erich and Vinyals, Oriol and Rae, Jack W. and Sifre, Laurent},
  year={2022},
  url={https://proceedings.neurips.cc/paper_files/paper/2022/file/c1e2faff6f588870935f114ebe04a3e5-Paper-Conference.pdf},
  booktitle={Advances in Neural Information Processing Systems},
  volume={35}
}

@inproceedings{penedo2024fineweb,
  title={The {FineWeb} Datasets: Decanting the Web for the Finest Text Data at Scale},
  author={Penedo, Guilherme and Kydl{\'i}{\v{c}}ek, Hynek and {Ben Allal}, Loubna and Lozhkov, Anton and Mitchell, Margaret and Raffel, Colin and {von Werra}, Leandro and Wolf, Thomas},
  booktitle={Advances in Neural Information Processing Systems},
  year={2024},
  url={https://proceedings.neurips.cc/paper_files/paper/2024/file/370df50ccfdf8bde18f8f9c2d9151bda-Paper-Datasets_and_Benchmarks_Track.pdf},
  volume={37}
}

@article{bussmann2024batchtopk,
  title={{BatchTopK} Sparse Autoencoders},
  author={Bussmann, Bart and Leask, Patrick and Nanda, Neel},
  journal={arXiv preprint arXiv:2412.06410},
  year={2024},
  url={https://arxiv.org/abs/2412.06410}
}

@article{rajamanoharan2024jumping,
  title={Jumping Ahead: Improving Reconstruction Fidelity with {JumpReLU} Sparse Autoencoders},
  author={Rajamanoharan, Senthooran and Lieberum, Tom and Sonnerat, Nicolas and Conmy, Arthur and Varma, Vikrant and Kram{\'a}r, J{\'a}nos and Nanda, Neel},
  journal={arXiv preprint arXiv:2407.14435},
  year={2024},
  url={https://arxiv.org/abs/2407.14435}
}

@inproceedings{chanin2025sparse,
  title={Sparse but Wrong: Incorrect {$L_0$} Leads to Incorrect Features in Sparse Autoencoders},
  author={Chanin, David and Garriga-Alonso, Adri{\`a}},
  year={2026},
  url={https://proceedings.mlr.press/v306/chanin26a.html},
  booktitle={Proceedings of the 43rd International Conference on Machine Learning},
  volume={306},
  pages={12889--12915},
  publisher={PMLR}
}

@inproceedings{clark2022unified,
  title={Unified Scaling Laws for Routed Language Models},
  author={Clark, Aidan and {de las Casas}, Diego and Guy, Aurelia and Mensch, Arthur and Paganini, Michela and Hoffmann, Jordan and Damoc, Bogdan and Hechtman, Blake and Cai, Trevor and Borgeaud, Sebastian and {van den Driessche}, George and Rutherford, Eliza and Hennigan, Tom and Johnson, Matthew and Millican, Katie and Cassirer, Albin and Jones, Chris and Buchatskaya, Elena and Budden, David and Sifre, Laurent and Osindero, Simon and Vinyals, Oriol and Rae, Jack and Elsen, Erich and Kavukcuoglu, Koray and Simonyan, Karen},
  booktitle={Proceedings of the 39th International Conference on Machine Learning},
  volume={162},
  pages={4057--4086},
  publisher={PMLR},
  year={2022},
  url={https://proceedings.mlr.press/v162/clark22a.html}
}

@inproceedings{chen2025revisiting,
  title={Revisiting Scaling Laws for Language Models: The Role of Data Quality and Training Strategies},
  author={Chen, Zhengyu and Wang, Siqi and Xiao, Teng and Wang, Yudong and Chen, Shiqi and Cai, Xunliang and He, Junxian and Wang, Jingang},
  booktitle={Proceedings of the 63rd Annual Meeting of the Association for Computational Linguistics (Volume 1: Long Papers)},
  pages={23881--23899},
  publisher={Association for Computational Linguistics},
  year={2025},
  doi={10.18653/v1/2025.acl-long.1163},
  url={https://aclanthology.org/2025.acl-long.1163/}
}

@inproceedings{allal2025smollm2,
  title={{SmolLM2}: When Smol Goes Big -- Data-Centric Training of a Fully Open Small Language Model},
  author={{Ben Allal}, Loubna and Lozhkov, Anton and Bakouch, Elie and {Mart{\'i}n Bl{\'a}zquez}, Gabriel and Penedo, Guilherme and Tunstall, Lewis and Marafioti, Andr{\'e}s and Kydl{\'i}{\v{c}}ek, Hynek and Srivastav, Vaibhav and Lochner, Joshua and Fourrier, Cl{\'e}mentine and Larcher, Hugo and Burtenshaw, Ben and Zhao, Haojun and Fahlgren, Caleb and Morlon, Mathieu and {Piqueres Lajar{\'i}n}, Agust{\'i}n and Nguyen, Xuan-Son and Zakka, Cyril and Raffel, Colin and {von Werra}, Leandro and Wolf, Thomas},
  year={2025},
  url={https://openreview.net/forum?id=3JiCl2A14H},
  booktitle={Second Conference on Language Modeling}
}
